\documentclass[letterpaper, 9 pt, conference]{ieeeconf}  % Comment this line out if you need a4paper

\IEEEoverridecommandlockouts                              % This command is only needed if 
\usepackage{epsfig} % for postscript graphics files
\usepackage{amsmath} % assumes amsmath package installed
\usepackage{amssymb}  % assumes amsmath package installed
\usepackage{bm}
\usepackage{dsfont}
\usepackage{color}
\usepackage{cite}
\usepackage{colortbl} 
\usepackage{diagbox}
\usepackage[absolute,overlay]{textpos}
\usepackage[linesnumbered,ruled]{algorithm2e}
\usepackage[normalem]{ulem} %to strike the words
\makeatletter
\let\NAT@parse\undefined
\makeatother
\usepackage{hyperref}
\hypersetup{pdfstartview=FitH,
            colorlinks=true,
            linkcolor=red,
            anchorcolor=blue,
            citecolor=black
            }
\usepackage{float}
\usepackage{booktabs}
\usepackage{arydshln}
\usepackage{bbm}
\usepackage[caption=false,font=footnotesize]{subfig}
\usepackage[dvipsnames]{xcolor}

\usepackage{soul}
\usepackage{multirow}
\usepackage{multicol}
\usepackage{graphicx}
\usepackage{subfloat}
\usepackage{amsmath}
\usepackage{amssymb}
\usepackage{booktabs}
\usepackage{algorithmic}
\usepackage[T1]{fontenc}
\usepackage[utf8]{inputenc}

\title{\LARGE \bf
Zero Shot Domain Adaptive Semantic Segmentation by Synthetic Data Generation and Progressive Adaptation
}

\author{Jun Luo$^{1}$, Zijing Zhao$^{1}$, Yang Liu$^{1}{^*}$% <-this % stops a space
\thanks{$^{*}$Corresponding author}% <-this % stops a space
\thanks{$^{1}$J. Luo, Z. Zhao, Y. Liu are with Wangxuan Institute of Computer Technology, Peking University, China. Email: {\tt\small \{luo\_jun, zijingzhao\}@stu.pku.edu.cn, yangliu@pku.edu.cn}}%
\thanks{This work was sponsored by Beijing Nova Program and supported by the grants from National Natural Science Foundation of China 62372014.}%
}
\begin{document}

\maketitle
\thispagestyle{empty}
\pagestyle{empty}

%%%%%%%%%%%%%%%%%%%%%%%%%%%%%%%%%%%%%%%%%%%%%%%%%%%%%%%%%%%%%%%%%%%%%%%%%%%%%%%%
\begin{abstract}

%\yang{The story is too big, you need to highlight you address OD or segmentation at the begining, rather than that late.}
Deep learning-based semantic segmentation models achieve impressive results yet remain limited in handling distribution shifts between training and test data. In this paper, we present SDGPA (Synthetic Data Generation and Progressive Adaptation), a novel method that tackles zero-shot domain adaptive semantic segmentation, in which no target images are available, but only a text description of the target domain's style is provided. 
%\zijing{We can mention the task name "domain adaptation for semantic segmentation" or "domain adaptive semantic segmentation", and explain the task briefly.}
To compensate for the lack of target domain training data, we utilize a pretrained off-the-shelf text-to-image diffusion model, which generates training images by transferring source domain images to target style. 
%\zijing{by transferring source domain images to target style}
Directly editing source domain images introduces noise that harms segmentation because the layout of source images cannot be precisely maintained.
% \yang{for each proposed solution, you need to tell them their motivation for the goal.}
% \zijing{Add a sentence like ``Editing source domain images introduce noise for segmentation because the layout of source images cannot be precisely maintained.''}
% \yang{please define method full name with abbre (SDGPA), here, as that is the keyword in your code link; try to repeat this method name few time throughout the paper as well.}
To address inaccurate layouts in synthetic data, we propose a method that crops the source image, edits small patches individually, and then merges them back together, which helps improve spatial precision.
% \yang{We propose cropping the source image and editing small patches individually, then merging them back together,which helps improve spatial precision.} 
% we propose cropping the source image and performing edits on small patches, which helps improve spatial precision. 
Recognizing the large domain gap, SDGPA constructs an augmented intermediate domain, leveraging easier adaptation subtasks to enable more stable model adaptation to the target domain. Additionally, to mitigate the impact of noise in synthetic data, we design a progressive adaptation strategy, ensuring robust learning throughout the training process. 
%\yang{too vague, how and why it works}.
%\zijing{do not mention "early stopping" because it is a commonly used optimization strategy that cannot be your contribution at all.}
%\zijing{emphasize only two parts of our method: synthetic data generation (with cropping) and progressive adaptation.}
Extensive experiments demonstrate that our method achieves state-of-the-art performance in zero-shot semantic segmentation. The code is available at https://github.com/ROUJINN/SDGPA
\end{abstract}

\section{Introduction}
%\yang{Given our method design is pretty simple, maybe you can read more therectical paper, to find math support for introducting such intermediate domain, and try to explain why this domain is useful (region-wise copy paste operation.)}

Semantic segmentation plays a crucial role in enabling scene comprehension for autonomous vehicles\cite{hofmarcher2019visual} and robotic applications\cite{liu2021light}. In the field of semantic segmentation, deep learning models\cite{ronneberger2015unet,lin2017refinenet,chen2018deeplabv3+} have demonstrated remarkable performance. These models, pretrained on annotated datasets, have achieved excellent results. However, when tested on data with distributions significantly different from the training data, their performance often degrades substantially. For instance, a segmentation model trained on clear weather cityviews may perform worse under snowy weather testing scenarios.
%\zijing{Since this is a robotics conference, we can mention the task name and explain the task in detail, maybe use an example like "for instance, a segmentation model trained on clear weather cityviews may perform worse under snowy weather testing scenarios." }
One potential solution is to increase the amount of training data from the target domain; however, annotating such data is often prohibitively time-consuming and resource-intensive, making it impractical in real-world scenarios.
%\yang{if you tell the story in this way. Readers will expect you will also conduct experiments on image classification at least,}

To address the aforementioned problem, researchers have proposed the task of unsupervised domain adaptation\cite{ganin2015uda}, aiming to leverage unlabeled target domain data to improve model performance on the target domain. Experiments have shown that their domain adaptation techniques effectively enhance model performance on the target domain\cite{hoyer2022daformer,zhao2023mrt}. Some researchers have explored more challenging scenarios where only a few\cite{fu2025crossfewshot} or a single image\cite{benigmim2023datum} from the target domain is available. However, these methods still require target domain images, which can be inconvenient in practice and makes them susceptible to bias from unrepresentative target domain samples.

This paper focuses on the scenario in which no target images are available, but only a text description of the target domain's style is provided, %\yang{style? How to describe a target domain by text, I guess just a style?}
known as zero-shot domain adaptive semantic segmentation\cite{peng2018zero}. In this line of research,  P{\O}DA\cite{fahes2023poda} utilizes text descriptions by leveraging CLIP\cite{radford2021clip} to obtain target-like features, and then trains the model on these features to drive its adaptation towards the target domain. 
% \yang{if they cannot access target domain data, how can they know target-like feature?}. 
However, this method requires careful feature selection and yields inferior 
% \yang{this word is not common, change to sth like `inferior?', replace this word throughou the paper}
results. 
%\yang{when you discuss other's limitation, try to make it concrete, you cannot just say not intuitive, what's your definition.}. 
ZoDi\cite{azuma2024zodi} attempts to use text-aligned Stable Diffusion\cite{rombach2022stablediff} model to generate target-like images but suffers from missing or inaccurate objects, resulting in a noticeable amount of noise in the synthesized data.
% s\zijing{This paragraph seems to be too detailed to appear in intro, maybe shorten the sentences and put detailed descriptions of the previous methods in related works.}
%\yang{do you have stats on how much missing?} 
It generally outperforms P{\O}DA, demonstrating significant potential in generating images for zero shot domain adaptation. However, it lacks effective mechanisms to reduce noise in synthetic data during either the data generation process or the model training phase, which ultimately limits its performance.
% \zijing{These contents will also be mentioned in related works, so here can be shorter.}

\begin{figure}[ht]
\vskip 0.2in
\begin{center}
\centerline{\includegraphics[width=\columnwidth]{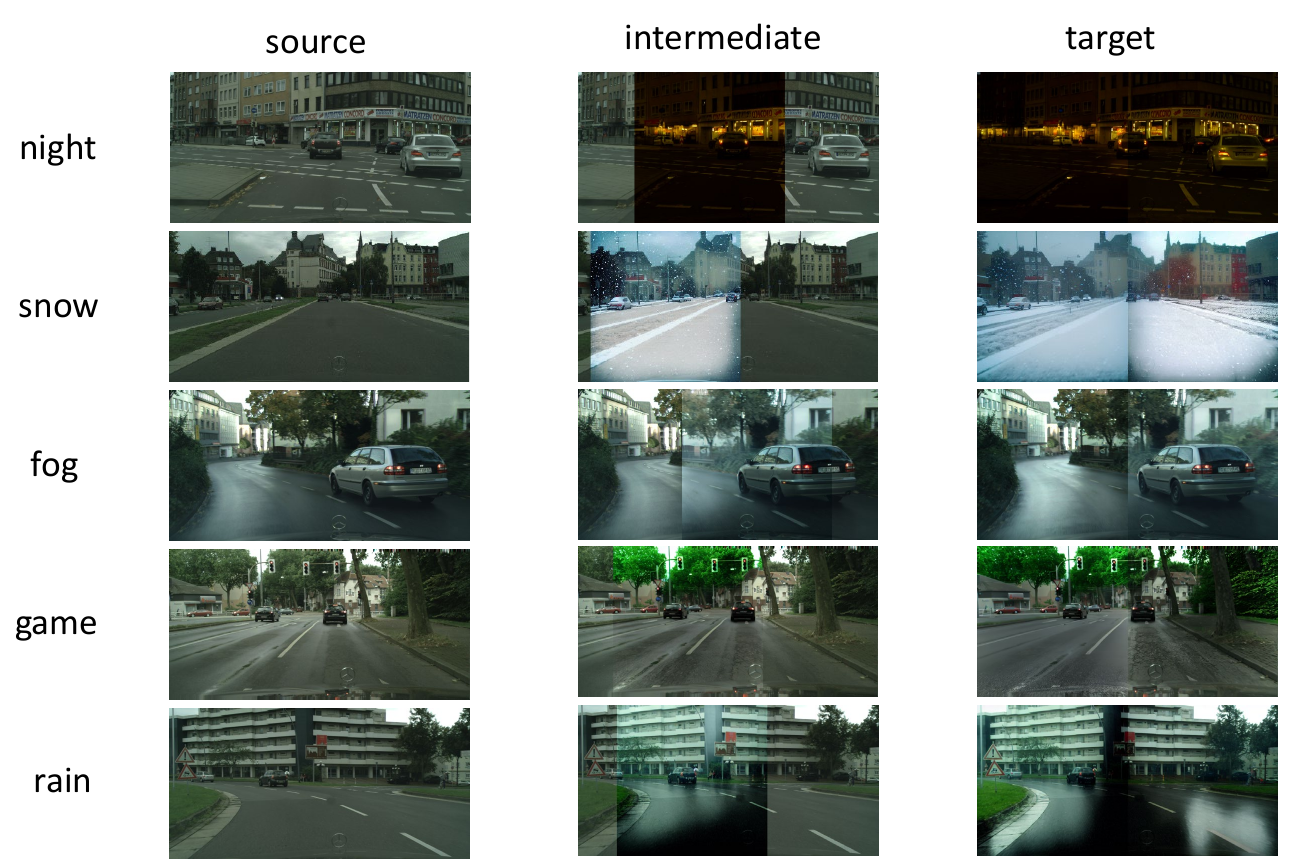}}
\caption{\textbf{An overview of the synthetic intermediate domain and the synthetic target domain we build}. Left: images from Cityscapes\cite{cordts2016cityscapes} dataset. Middle: synthetic intermediate domain by editing a patch of source image. Right: synthetic target domain by entirely transforming source image}
\label{tastefig}
\end{center}
\vskip -0.2in
\end{figure}

To this end, we propose SDGPA, a novel pipeline for zero-shot domain adaptation with the help of a recently proposed image editing diffusion model\cite{brooks2023instructpix2pix}. Recognizing that the editing model can effectively preserve image layouts when operating on small patches where it is trained, we propose cropping the source image and performing edits on localized patches. 
%\yang{you did not provide the motivation on this copy transform paste operation.}
Such an approach provides greater flexibility in our design. Specifically, we construct an augmented intermediate domain by 
% \yang{based on your following description in this paragraph, I cannot see what's idfferent between intermediate and tartget iamge; try to set contrastive words to guide reader, like intermediate , one random crop patch?  target: all patches transfer one by one ,and merge all? plz revise.}
randomly cropping a patch from the source image, and then transferring the patch's style with image editing diffusion model
% \yang{transfer the style?}
and pasting it back into its original position. The intermediate domain serves as a bridge between the source and target domain, thus avoiding direct mapping between two significantly different distributions\cite{hsu2020progressive} and makes it easier for the models to adapt to the target domain. 
% \zijing{This expression looks like we take other method's idea. We can say ``We propose to utilize an intermediate domain ..., inspired by xxx.''}
We describe the intermediate domain as augmented because the random cropping operation introduces diverse, even disharmonious images, enabling the model to learn more robust features, and the crop-and-paste operation\cite{ghiasi2021copypaste} is a highly effective augmentation for segmentation task.
%\zijing{Try not to say directly that our operation has already been proven. It weakens our novalty. Just say ``We crop ... and paste ... which is a highly effective augmentation for segmentation task [19].''}
By deterministically dividing the source image into multiple non-overlapping patches and then editing all patches and merging them, the source image can be entirely transformed into a target-like image, effectively creating a synthetic target domain. The generated images are shown in Figure \ref{tastefig}. To better utilize these two domains, we further propose a progressive adaptation strategy. Specifically, we first train the model on both the source and synthetic intermediate domains, then train it on both the synthetic intermediate and synthetic target domains. To avoid overfitting to noisy data, we early stop training in the second step. 
%\yang{early stop is a implementation detail, not the key}
This simple yet effective training approach allows the model to better adapt to the synthetic target domain while being less affected by the noise introduced by synthetic images. 
% \yang{Highly suggest you read some theoretical paper on introducing intermediate domain to bridge the gap }

The contributions of this paper can be summarized as follows: a) We present SDGPA, a novel method suitable for diffusion model assisted zero shot domain adaptative semantic segmentation. b) We effectively construct synthetic augmented intermediate domains and synthetic target domains and introduce a progressive adaptation strategy that efficiently utilizes the synthetic intermediate and target domains. c) Our zero-shot method for training semantic segmentation models achieves state-of-the-art results on multiple datasets, proving the effectiveness of our method.
\section{Related Works}

\subsection{Gradual Domain Adaptation}

% Unsupervised Domain Adaptation addresses the domain gap by taking full advantage of unannotated target domain images. Most of them employ a self-training framework\cite{tarvainen2017meanteacher}, where a teacher model is utilized to generate pseudo-labels for images in the target domain and a student model is trained using these pseudo-labels along with the ground truth from the source domain. Various techniques, such as adversarial training\cite{chen2018domainadversial}, pseudo-label refinery\cite{wang2021pseudolabelrefinery}, weak-strong augmentation\cite{liu2021unbiasedteacher}, are widely adopted in this process.

Unsupervised Domain Adaptation addresses the domain gap by taking full advantage of unannotated target domain images. Facing the large domain gap between the source and target domains, motivated by the classic divide-and-conquer strategy, Gradual Domain Adaptation\cite{chen2021gradual,he2024gradual,shi2024adversarial} (GDA) leverages intermediate data to address significant distribution shifts by solving intermediate adaptation problems and then combining these solutions to tackle the original domain shift. Theoretical analyses on generalization error bounds and optimal paths, along with extensive experiments, have demonstrated the effectiveness of this approach. However, GDA assumes the availability of unlabeled intermediate data, which is often impractical in real-world scenarios. % \yang{Note you cannot use reference if as the sentnece subject, change to most relevant work [18][25]?; please fix other similar issues across the paper.}
Our work and several relevant works\cite{gopalan2011domain,hsu2020progressive} share the same core idea with GDA. He \textit{et al.}\cite{gopalan2011domain} align intermediate feature representations, while Hsu \textit{et al.}'s work\cite{hsu2020progressive} and our work construct a synthetic intermediate domain by generating images. The key difference between our work and Hsu \textit{et al.}'s work\cite{hsu2020progressive} lies in our approach of locally editing random patches from the source image to form the intermediate domain.
%\zijing{The grammar of this sentence seems to be strange, maybe use GPT to check the grammar and wording.}
% \zijing{We should not only say the difference in operation, but also the motivation: what is the fact that [18] doesn't consider, so it is worse than ours. 这个说起来有点太麻烦，因为他首先跟我们不是同一个setting下的，其次他为了解决noise主要用的是给图片加权重，但是权重又需要通过他里面的一个target domain discriminator来获得，which is 偏离我们文章内容的东西}
This approach naturally introduces data augmentation while preserving more precise layouts.

\subsection{Zero Shot Domain Adaptation}

Under the zero shot setting, where no target images are available during training, a pioneering work\cite{peng2018zero} utilizes task-irrelevant dual-domain pairs to learn applicable representations both in source and target domain. Recently, P{\O}DA\cite{fahes2023poda} leverages a pretrained contrastive vision-language model (CLIP\cite{radford2021clip}) to transfer the features in source domain to target domain with text prompt guidance, then finetunes perception models on transferred features with original annotation. However, it assumes a specific backbone and yields inferior
% \yang{change this word}
results. Owing to the remarkable ability of recently advanced diffusion models in generating diverse and high-quality images, several studies\cite{azuma2024zodi,shen2024controluda} have attempted to leverage Stable Diffusion and ControlNet\cite{zhang2023controlnet} to produce images that mimic the style of the target domain. These generated images, along with the original real images, are then used to train models. 
% However, these methods often face challenges such as insufficient style transfer and imprecise object generation, leading to inconsistencies between synthetic target domain images and their annotations. 
However, we notice that two key aspects have been overlooked by previous works: firstly, it is easier for pretrained diffusion models to transfer style and maintain objects at a certain resolution, and secondly, one-stage adaptation fails to fully utilize the generated data due to the large domain gap.
%\zijing{However, we find out that pretrained diffusion models only work well on fixed resolution, and the one-stage adaptation can not make fully use of the generated data due to the large domain gap, which has not been considered by previous works. (Change the expression of these motivations and fuse it into the introduction of related works.)}
SDGPA incorporates an intermediate domain construction approach and a progressive adaptation strategy and has effectively mitigated the aforementioned issues, ultimately achieving superior performance in the unseen target domain.

\subsection{Data Augmentation}

Data augmentation plays an important role in the field of computer vision\cite{yang2022augmentationsurvey}. As an effective way to improve the sufficiency and diversity of training data, it helps prevent overfitting and enables models to learn robust representations. Standard data augmentation techniques include random crop\cite{krizhevsky2012alexnet}, color jittering\cite{szegedy2015goingdeeper}, random flipping, and so on. A novel data augmentation method for instance segmentation is Copy-Paste augmentation\cite{ghiasi2021copypaste}. It employs a simple mechanism: randomly selecting objects and pasting them at random locations on the target image, while correspondingly adjusting the annotations. Although this approach generates visually inconsistent images, it proves highly effective for training instance segmentation models. Our method shares the same spirit in producing such inconsistent images, leveraging their utility for model training.

\subsection{Diffusion Model}

Diffusion models, first introduced by \cite{sohl2015deep,ho2020ddpm}, have revolutionized the field of image generation\cite{dhariwal2021diffbeatsgan}. Beyond image generation, diffusion models have expanded into diverse tasks, including Style Transfer\cite{wang2024instantstyle,chung2024styleid}, Layout Control\cite{zhang2023controlnet}, Personalized Generation\cite{ruiz2023dreambooth,gal2022textualinversion}, and image editing\cite{brooks2023instructpix2pix,kawar2023imagic}. Image editing models show their excellent ability to follow input edit prompt and maintain input image content. For our work, we choose InstructPix2Pix\cite{brooks2023instructpix2pix} to help us accomplish zero shot domain adaptation.
% \zijing{Maybe mention why we choose InstructPix2Pix. 这个感觉再夸他就有点过了，我们要强调我们的东西}

% \zijing{
% The length of related work is a little bit too long. Consider shorten the sentences, making each subsection 1 paragraph.
% Sec 2.2 which tells the difference between our methods and direct competitors could be appropriately detailed.
% }

%\yang{For related work, the purpose of each sub-section is to elaborate your novelty. You can discuss some most relevant work, but the end, you need few sentence or a even a small paragraph to discuss how you distinct from them. To me you do not propose sth new, for image generation, maybe no need for such a long paragraph.}
\section{Method}
\label{sec:method}

% \zijing{
% I recommand adding a subsection to describe the task setting as 'problem formulation'.
% Define the training data, testing data, available text style description, and emphasize that our used text description is dataset level, and finally conclude that this setting is challenging.
% }

\subsection{Problem Formulation}
\label{problemform}

We focus on the problem of \textit{zero-shot domain adaptive semantic segmentation}, as proposed in \cite{fahes2023poda}. Let \( D_s = \{(x_s, y_s) \mid x_s \in \mathbb{R}^{H \times W \times 3}, y_s \in \{0, 1\}^{H \times W \times L}\} \) denote the source domain dataset, where \( x_s \) represents the input RGB images and \( y_s \) denotes the corresponding pixel-wise semantic labels. Additionally, a high-level description of the target domain (e.g., ``rainy'') is provided. The goal is to train a $L$-class segmentation model $M$ that can generalize effectively to an unseen target domain \( D_t = \{(x_t, y_t) \mid x_t \in \mathbb{R}^{H \times W \times 3}, y_t \in \{0, 1\}^{H \times W \times L}\} \), thereby improving its segmentation performance under the target domain's specific conditions, without access to either images $x_t$ or annotations $y_t$ from the target domain during training.

\subsection{Overview}

\begin{figure*}[ht]
\vskip 0.2in
\begin{center}
\centerline{\includegraphics[width=2\columnwidth]{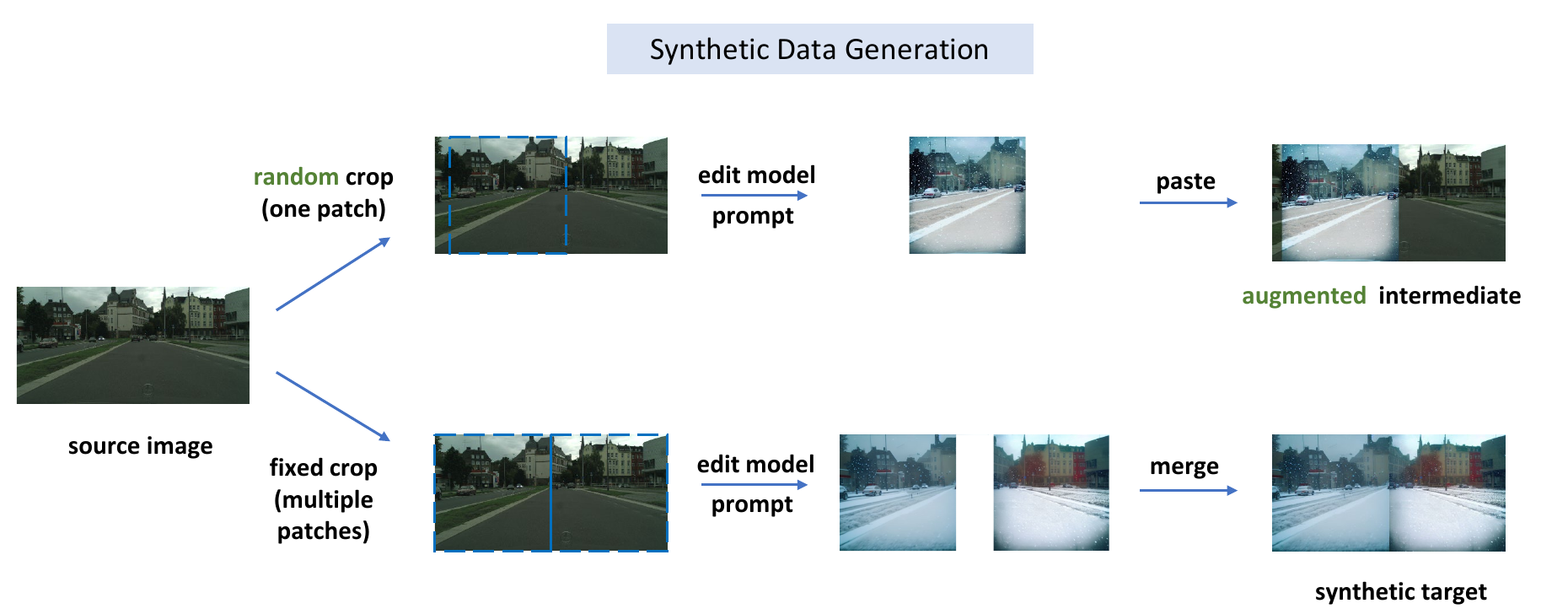}}
%\yang{In such an image, you might want to highlight the point you propose, rather than base model  you use, especially those you do not add anything new on top.}
\caption{\textbf{Synthetic Data Generation.} To build synthetic intermediate domain, we randomly crop a fixed resolution patch from the source image. Then we use a diffusion-based image editing model\cite{brooks2023instructpix2pix} along with an edit prompt to transfer the cropped patch into our desired style and paste it back onto the source image. For synthetic target domain, we deterministically divide the source image into multiple non-overlapping patches and then edit all patches and merge them to ensure a complete transfer of the source image. We denote the two approaches as random crop and fixed crop. 
% \yang{change the text in fig, intermediate, single random crop? target: all crops?; change the word concat to merge makes it sound more general.fix image content as well as caption and in-line description, to make them all consistent.}
}
\label{methodfig}
\end{center}
\vskip -0.2in
\end{figure*}

Our method SDGPA consists of two key stages: \textit{Synthetic Data Generation} and \textit{Progressive Adaptation}. The zero shot domain adaptation setting is inherently challenging due to the absence of labeled target domain data. To address this, we leverage pretrained image editing diffusion models to generate target-style images while preserving the semantic layout of the source domain. As illustrated in Figure \ref{methodfig}, in the first stage, we construct a synthetic intermediate domain and a synthetic target domain, providing sufficient and diverse data for the segmentation model to adapt to unseen target domain. Owing to our cropping-based design, the semantic layouts of these synthetic images remain highly consistent with the source images, as demonstrated in Figure \ref{tastefig}. Despite this consistency, the generated data still exhibits a non-negligible domain gap compared to the real target domain. To mitigate this, we introduce a progressive adaptation strategy in the second stage, as shown in Algorithm \ref{alg1}. This approach gradually adapts the segmentation model to the synthetic target domain, stabilizing the training process and preventing overfitting. By combining these two stages, our method effectively bridges the domain gap and enhances the model's generalization capability in zero-shot settings.

% \zijing{
% Add motivation to the overview.
% Zero-shot setting is challenging, so we employ pretrained img2img diffusion model to generate target-style data while keeping the segmentation layout.
% The diffusion model works well only on fixed resolution where is it pretrained, so we design synthetic data generation pipeline with combined cropping.
% The generated data, though with accurate annotation from source domain, has relatively large domain gap, so we design progressive adaptation.
% }

\subsection{Synthetic Data Generation}
\label{method_sdg}

Recent advancements in image-to-image diffusion models, such as ControlNet \cite{zhang2023controlnet} and InstructPix2Pix \cite{brooks2023instructpix2pix}, have demonstrated remarkable capabilities in transferring image styles while preserving their underlying layouts. These models provide a promising foundation for generating synthetic images tailored to zero-shot domain adaptive semantic segmentation tasks. Previous works, such as ZoDi \cite{azuma2024zodi}, faced challenges in constructing synthetic domains. These challenges included the loss of small objects, the generation of incorrect objects, and the loss of image content, all of which introduce significant noise and degrade model performance. To address these issues, we aim to leverage diffusion models to build a domain that maintains more precise alignment with the source annotations for segmentation model training. Motivated by the powerful image editing capabilities of editing models\cite{brooks2023instructpix2pix} at the fixed resolution where it is trained and the intermediate synthetic domain in \cite{hsu2020progressive}, we propose a cropping-based method for both data augmentation and image style transfer. 

% \zijing{Change the expression of this paragraph. 
% Thanks to the development of img2img diffusion models like ControlNet and Instruct Pix2Pix, we are able to transfer style of the source images while keeping its layout.
% In this way, we can create synthetic images for zero-shot domain adaptive semantic segmentation.
% Previous methods (Zodi) ignores the limitation of img2img models that it only works well on the pretrained resolution.
% To overcome this, we design synthetic data generation pipeline with combined cropping to make fully use of pretrained diffusion models.
% }

First, to reduce computational costs and capture richer semantic information during image editing, we first resize the source domain images such that the shorter edge of the image is the same as the pre-trained size of the diffusion model while maintaining the aspect ratio. For the augmented synthetic intermediate domain, we randomly crop a fixed resolution patch from the source image and use image editing model along with an editing prompt (e.g., ``make it rain'') to transform the cropped patch into the desired target domain. The edited patch is then pasted back into the source image at its original location. Although this intermediate domain may appear visually inconsistent, the annotations remain sufficiently accurate for the segmentation model, ultimately enabling the model to learn more robust feature representations through this augmented intermediate domain. 

To further push the model's capability toward the target domain, we aim to fully transform the source image into an image resembling the unseen target domain while preserving the layout and avoiding missing objects. To achieve this, we deterministically divide the source image into multiple non-overlapping patches.
%\yang{try to revise description, to make it sound more general, like crop into multiple crops, } 
These patches are individually edited by the editing model and then merged together to form the synthetic target domain image. This approach ensures both style transfer and layout consistency, addressing the limitations of previous methods.

% \zijing{
% We can rearrange the two paragraphs as:
% We propose two strategies of generating target-style images, random cropping and fixed cropping.
% Random cropping ....(describe the operation). It generates augmented intermediate domain images in which some part of the image remains source style.
% Fixed cropping ...(describe the operation).
% It generates synthetic target domain images in which the whole image in transferred to target style with better quality, making full use of the pretrained diffusion.
% }

\subsection{Progressive Model Adaptation}

% \zijing{I recommand using algorithm flow to describe this section instead of the image of the method.}

% \begin{algorithm}
% \caption{Progressive Adaptation}
% \label{alg:counterfactual}
% \SetKwInOut{Input}{Input}
% \SetKwInOut{Output}{Output}
% \begin{algorithmic}[1]
% \Input{Source domain data $D_s = \{(x_s, y_s) \mid x_s \in \mathbb{R}^{H \times W \times 3}, y_s \in \{0, 1\}^{H \times W \times L}\}$, synthetic intermediate data $D_{si} = \{x_{si}|x_{si}\in \mathbb{R}^{H \times W \times 3} \}$, synthetic target data $D_{st} = \{x_{st}|x_{st}\in \mathbb{R}^{H \times W \times 3} \}$.}
% \Output{a semantic segmentation model M}

% \end{algorithmic}
% \end{algorithm}

\begin{algorithm}
\caption{Progressive Adaptation}
\label{alg1}
\SetKwInOut{Input}{Input}
\SetKwInOut{Output}{Output}
\Input{Source domain data $D_s = \{(x_s, y_s)\}$, synthetic intermediate data $D_{si} = \{x_{si}\}$, synthetic target data $D_{st} = \{x_{st}\}$, an initialized semantic segmentation model $M$, total epochs $N$, early stop ratio $\alpha<1$.}
\Output{Model $M$ that performs well on the unseen target domain.}

\For{epoch\_number $\gets 1$ to $N$}{
    \For{each batch of paired $(x_s, x_{si})$}{
        Compute the loss $L_1$ as sum of cross-entropy losses: \\
        \quad $L_1 \gets \text{CE}(x_s, y_s) + \text{CE}(x_{si}, y_s)$ \\
        Update $M$ by taking a gradient step on $L_1$.\\
    }
}

\For{epoch\_number $\gets 1$ to $N$}{
    \For{each batch of paired $(x_{si}, x_{st})$}{
        Compute the loss $L_2$ as sum of cross-entropy losses: \\
        \quad $L_2 \gets \text{CE}(x_{si}, y_s) + \text{CE}(x_{st}, y_s)$ \\
        Update $M$ by taking a gradient step on $L_2$.\\
    }
    \If{epoch\_number $== \alpha N$}{
        Break the loop.\\
    }
}
\Return $M$\;
\end{algorithm}

A critical challenge remains in effectively leveraging our synthetic data for zero-shot domain adaptive semantic segmentation. Unlike unsupervised domain adaptation (UDA) scenarios, where target images strictly follow the target domain's distribution but lack annotations, our synthetic target domain inherently provides relatively accurate annotations while not fully conforming to the target domain's distribution. Ideally, fine-tuning a source-pretrained segmentation model solely on the synthetic target domain should yield satisfactory performance. However, the presence of noise significantly impacts the results, limiting the model's performance on the unseen target domain. 
ZoDi\cite{azuma2024zodi} proposed incorporating a feature similarity loss during training, but this approach did not address the noise and did not yield significant performance improvements. 
% \zijing{In this sentence we need some reason why the feature similarity loss cannot address the issue.}

In the early experiments, we observed that fine-tuning a source-pretrained model solely on the synthetic dataset consistently underperforms compared to joint training on both the source and synthetic datasets when evaluated on the unseen target domain. We hypothesize that during adaptation, timely supervision from the source domain's real images and accurate annotations is crucial, as it mitigates the impact of noise in the synthetic dataset. However, our objective remains to improve performance on the unseen target domain. To this end, we propose a Progressive Model Adaptation strategy, as shown in Algorithm \ref{alg1}. First, we train the model $M$ with both source domain and synthetic intermediate domain for $N$ epochs (Line 1-7), Then we continue to train $M$ with synthetic intermediate domain and synthetic target domain (Line 8-17).
%\zijing{Add a few sentences to describe the algorithm, not just let the readers read the algorithm flow.}
% \zijing{We cannot express the domain as a loss.
% We need to formulate the loss function. For example, $\mathcal{L}_s = \mathcal{L}_{seg}(x_s, a_s) + \mathcal{L}_{seg}(x_m, a_s)$, where $\mathcal{L}_{seg}$ represents the original loss of training the segmentation model, $x_s$ and $x_m$ denote source domain image and intermediate domain image respectively, and $a_s$ denotes source domain annotation.} 这里上面的语句中说到了是Denoting the cross-entropy loss as ...
% \zijing{If you think it is important, split another paragraph to describe early stopping.
% First describe the fact that all existing domain adaptive segmentation methods train the model for a fixed number of epochs.
% We discover that early stopping is a simle yet effective strategy to avoid overfitting on this task.
% The reason is that in the first stage ..., second stage ...
% Don't use number of 65 and 100 epochs and 3/4 directly in the method section.
% }
To mitigate the impact of noise during training, we employ early stopping at the $\alpha N$th epoch to avoid overfitting (Line 14-16).
% \zijing{To avoid the impact of noise ..., we employ early stopping ...}
We analyze its success based on three key factors: a) At each stage, the model benefits from the augmented synthetic intermediate images. b) During the first stage, the model is primarily guided by high-quality source images, which facilitate learning robust representations and enable gradual adaptation to the unseen target domain. In the second stage, a greater proportion of the supervisory signal is derived from synthetic images that are more similar to the unseen target domain, thereby further driving the model's adaptation. Meanwhile, residual supervision from source images helps to counteract the potential noise introduced by synthetic data.
%\zijing{the expression "more supervision comes from ..." seems to be not professional enough, maybe use GPT to refine the expression.}
c) As elucidated in \cite{arpit2017closer} regarding the memorization effect - wherein deep neural networks initially memorize and fit the majority (clean) patterns and subsequently overfitting to minority (noisy) patterns - thus the early stopping strategy\cite{tanaka2018joint} can prevent the model from overfitting to noisy data.

\section{Experiments}

\subsection{Experimental Design}

Following ZoDi\cite{azuma2024zodi} and P{\O}DA\cite{fahes2023poda}, we conduct experiments on zero shot domain adaptative semantic segmentation, as described in Section \ref{problemform}. 
%\zijing{This sentence seems to have no information, maybe you want to say "we conduct experiments on zero-shot DASS task setting as described in Section 3.1"}

For each setting, first, we use InstructPix2Pix\cite{brooks2023instructpix2pix} and the method described in Section \ref{method_sdg} to obtain synthetic intermediate domain data $D_{si}$ and synthetic target domain data $D_{st}$. We then train a DeepLabv3+\cite{chen2018deeplabv3+} model $M$ using these data and Algorithm \ref{alg1}. Finally, we evaluate the performance of the obtained model. 

\subsection{Implementation Details}

\subsubsection{Datasets}
We use CityScapes\cite{cordts2016cityscapes} dataset as the source domain. It contains 2975 training images and 500 validation images, which were collected from 50 cities in the daytime. We consider several zero shot domain adaptation settings: day $\to$ night, clear $\to$ snow, clear $\to$ rain, clear $\to$ fog. For these settings, we use ACDC\cite{sakaridis2021acdc} dataset. ACDC comprises a large set of 4006 images which are evenly distributed between fog, nighttime, rain, and snow weather conditions. Under real $\to$ game setting, we use GTA5\cite{richter2016gta5} dataset, whose images are rendered using the open-world video game Grand Theft Auto 5 and are all from the car perspective in the streets. We evaluate our method's performance using the validation set of ACDC and the same random subset of 1000 images from GTA5 as ZoDi and P{\O}DA use.

\subsubsection{Data Generation}

We use InstructPix2Pix\cite{brooks2023instructpix2pix}, which is an image editing model built upon Stable Diffusion v1-5\cite{rombach2022stablediff}. 
%\zijing{to do something or as a base model to do something}
To obtain data that is effective for training, we need to carefully set hyperparameters to prevent insufficient style transfer or excessive damage to the original image content. Empirically, we observe that using edit prompts included in InstructPix2Pix's training set yields better performance. We summarize our hyperparameter design for different settings in Table \ref{paramtab}. 
% \zijing{
% We can mention here that our method only need simple text prompt for a target domain.
% }
For a given setting, the synthetic intermediate domain and synthetic target domain share the same hyperparameters. Similar to ZoDi, the number of images in each of our synthetic domains is the same as in the Cityscapes training set. Each image corresponds one-to-one with the original image.

% All of the images' resolutions are summarized in Table \ref{resolutable}. 
% \input{tabel/resolutab}

\subsubsection{Training}

Following ZoDi, we use the DeepLabv3+\cite{chen2018deeplabv3+} architecture with the backbone using ResNet-50\cite{he2016resnet} as the base semantic segmentation model. We initialize the backbone with ImageNet-1K\cite{deng2009imagenet} pretrained weights. During training, we resize all input images to 512$\times$1024 resolution. We train the model on random 384$\times$768 crops with batch size 4. We apply standard color jittering and horizontal flip to crops. We use a polynomial learning rate schedule with an initial learning rate $10^{-3}$, power $0.9$. SGD with momentum 0.9 and weight decay $10^{-4}$ for optimization. We choose $N = 100$, and $\alpha = 0.65$.

\subsubsection{Evaluation}

We use Mean Intersection over Union (mIoU) to evaluate our segmentation model's performance. During evaluation, we follow the same scheme as in training to resize all input images to 1024$\times$512 resolution. We report the mIoU over three models trained with different seeds in the second stage of Progressive Model Adaptation.

\begin{table*}[t]
\caption{\textbf{Results of Semantic Segmentation} under different conditions(in mIoU). \#Tar. represents the number of images from target domain that are used during training. }
\label{maintab}
\vspace{-8pt}
\begin{center}
\begin{small}

\begin{tabular}{llccccc}
\toprule
    Method & \#Tar.  & Day→Night & Clear→Snow & Clear→Rain & Clear→Fog & Real→Game \\
\midrule
    Source-only & -         & 20.7±1.0 & 40.8±0.3 & 39.4±0.3 & 54.5±1.0 & 40.1±0.7 \\
    DAFormer\cite{hoyer2022daformer} & 400     & 30.3±0.3 & 44.5±1.3 & 41.0±2.2 & 58.2±0.3 & 47.9±0.6 \\
\midrule
    DATUM\cite{benigmim2023datum}+DAFormer & one        & 19.9±0.6 & 43.0±0.1 & 39.0±1.5 & 60.3±0.4 & 44.7±0.7 \\
\midrule
    PØDA\cite{fahes2023poda} & zero      & 25.0±0.5 & 43.9±0.5 & 42.3±0.6 & 49.0±0.9 & 41.1±0.5 \\
    ZoDi\cite{azuma2024zodi}  & zero       & 24.7±0.2 & 45.6±1.6 & 47.0±1.2 & 56.1±1.7 & 40.5±0.2 \\
    SDGPA (ours)  & zero      & \textbf{26.9}±0.8 & \textbf{47.4}±0.7 & \textbf{48.6}±0.8 & \textbf{58.8}±0.7 & \textbf{43.4}±0.4 \\
\bottomrule
\end{tabular}

\end{small}
\end{center}
\vskip -0.1in
\end{table*}

\begin{table*}[t]
\caption{Hyperparameter for InstructPix2Pix to edit image to different domain}
\label{paramtab}
\vspace{-8pt}
\begin{center}
\begin{small}

\begin{tabular}{llcc}
\toprule
          & Edit prompt & Text guidance scale & Image guidance scale \\
\midrule
    Snow  & \textnormal{What would it look like if it were snowing?} & 15    & 1.1 \\
    Fog   & \textnormal{Make it foggy} & 7.0     & 2.0 \\
    Night & \textnormal{Make it evening} & 20    & 1.7 \\
    Rain  & \textnormal{Make it rain} & 20    & 2.0 \\
    Game  & \textnormal{Have it be a 3d game} & 12    & 2.0 \\
\bottomrule
\end{tabular}

\end{small}
\end{center}
\vskip -0.1in
\end{table*}

\subsection{Main Results}

We summarize our results in Table \ref{maintab}. We compare our method's performance with previous zero shot state-of-the-art ZoDi\cite{azuma2024zodi} and P{\O}DA\cite{fahes2023poda}. We also compare with several methods under less constrained setting. DAFormer\cite{hoyer2022daformer} is a strong baseline for unsupervised domain adaptative semantic segmentation, which requires using all images from the target domain. DATUM\cite{benigmim2023datum} is a one shot method that leverages crops from one target image to finetune a pre-trained diffusion model and generates unannotated target domain image, thus can be used with DAFormer. 

\subsubsection{Comparison to Zero Shot Methods}

\begin{figure}[ht]
\vskip 0.2in
\begin{center}
\centerline{\includegraphics[width=\columnwidth]{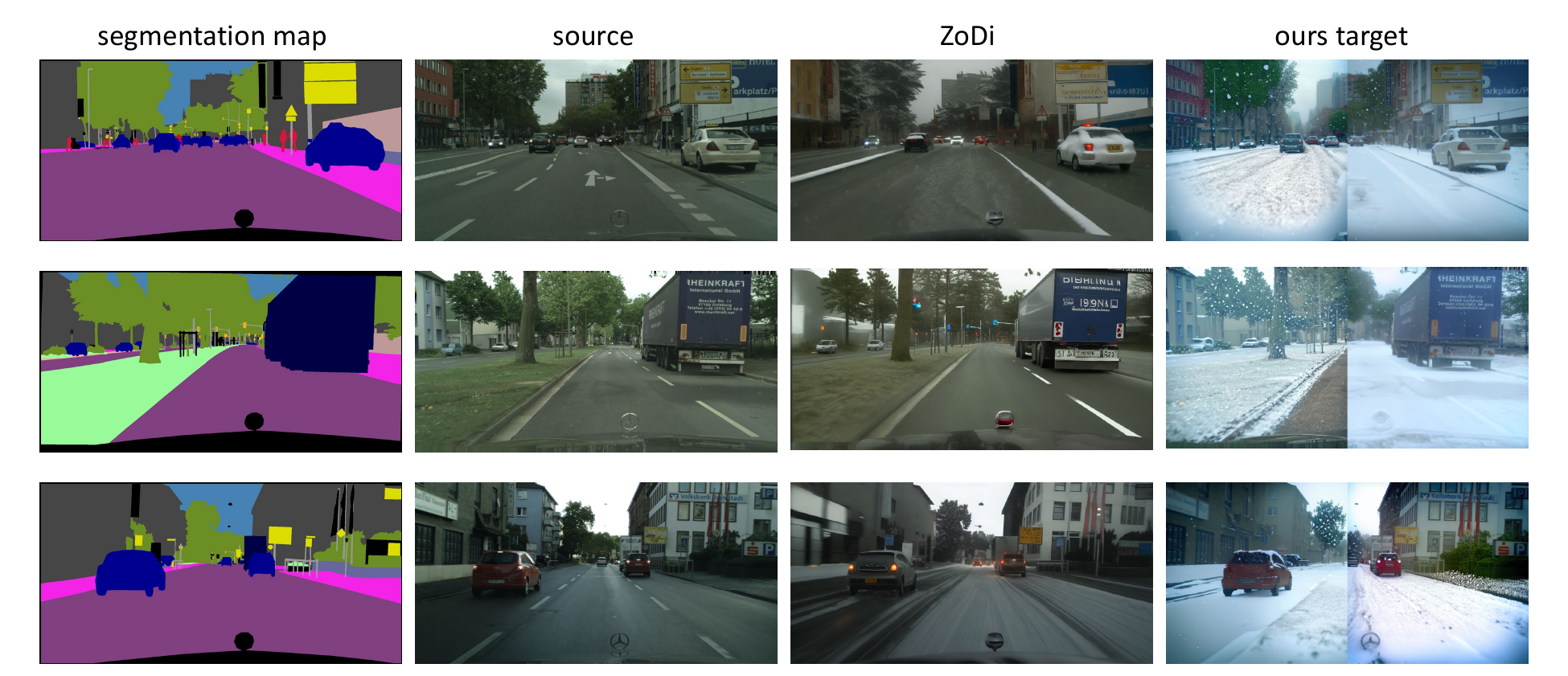}}
\caption{\textbf{A qualitative comparison of our method with ZoDi}. ZoDi cannot  effectively add snow (row 1 and 2), or cannot maintain objects accurately. In row 2, it fails to generate a house on the left, and in row 3 it misses a small car. Although our method still suffers from the loss of small objects (row 1), our method overall demonstrates better consistency with the source annotations and transfers style more effectively.}
\label{comparetozodifig}
\end{center}
\vskip -0.2in
\end{figure}

% \yang{Clear state As shown in which table? You need to guide readers to look for these numbers; If you re-run exp, remember to revise all numbers discussion in exp to make it aligned.}
As shown in Table \ref{maintab}, our method demonstrated substantial improvements over the source-only segmentation model in all settings: +6.2 for day$\to$night, +6.6 for clear$\to$snow, +9.2 for clear$\to$rain, +4.3 for clear$\to$fog, and +3.3 for real$\to$game. Furthermore, we have successfully achieved state-of-the-art zero-shot results across all datasets. This includes settings where the diffusion-based ZoDi did not surpass P{\O}DA, such as day$\to$night and real$\to$game, with improvements of +1.9 and +2.3, respectively. In settings where ZoDi outperformed P{\O}DA,(clear$\to$snow, clear$\to$rain, and clear$\to$fog), we still achieved significant improvements over ZoDi, with gains of +1.8, +1.6, and +2.7, respectively. 
%\zijing{First say we outperform source only by large marigins, then say this improvement achieve SOTA compared with previous methods.}
Our state-of-the-art performance across all datasets highlights the effectiveness of the proposed approach. We attribute this success to three key factors: a) the construction of an augmented intermediate domain, which provides robust features and enables progressive adaptation. b) the proposed progressive training strategy allows the model to better adapt to the synthetic target domain while being less affected by the noise introduced by synthetic images. c) operating on image crops, which ensures the accuracy of the synthesized image layouts. A qualitative comparison under the clear$\to$ snow setting is illustrated in Figure \ref{comparetozodifig}. A quantitative comparison is shown in Table \ref{classtab}. Our method achieves superior performance over ZoDi across multiple categories, including large objects such as buildings, walls, and buses, as well as small objects like traffic signs and poles. However, it is noteworthy that our performance on certain objects, such as cars, trucks, and riders, falls short compared to ZoDi. We attribute this to the editing model altering too much of their appearance or blurring them. Additionally, for some classes like the sky, bicycles, and motorcycles, the source-only approach yields the best results, underscoring the importance of incorporating source images during training.

\begin{table}[ht]
\caption{We randomly choose a seed and report quantitative comparison of different classes' IoU on unseen target domain between different methods under the clear$\to$ snow setting.}
\label{classtab}
\vspace{-8pt}
\begin{center}
\begin{small}

\begin{tabular}{lccc}
\toprule
    Class    & source-only & ZoDi\footnotemark & SDGPA(ours) \\
\midrule
    road  & 76.1  & 80.9  & \textbf{80.9} \\
    sidewalk & 37.6  & \textbf{46.2}  & 44.6 \\
    building & 68.2  & 67.1  & \textbf{71.2} \\
    wall  & 32.9  & 30.8  & \textbf{36.0} \\
    fence & 29.6  & 29.1  & \textbf{31.2} \\
    pole  & 24.6  & 27.1  & \textbf{29.3} \\
    traffic light & 50.5  & 50.2  & \textbf{58.1} \\
    traffic sign & 45.7  & 47.1  & \textbf{51.5} \\
    vegetation & 80.1  & 79.5  & \textbf{80.7} \\
    terrain & 8.8   & \textbf{12.6}  & 9.1 \\
    sky   & \textbf{86.5}  & 85.9  & 86.3 \\
    person & 39.6  & 50.3  & \textbf{53.7} \\
    rider & 2.8   & \textbf{7.8}   & 0.2 \\
    car   & 60.9  & \textbf{75.3}  & 71.6 \\
    truck & 42.4  & \textbf{70.3}  & 51.5 \\
    bus   & 19.2  & 25.6  & \textbf{29.0} \\
    train & 53.9  & 61.5  & \textbf{67.9} \\
    motorcycle & \textbf{28.5}  & 16.6  & 21.1 \\
    bicycle & \textbf{41.5}  & 22.2  & 34.0 \\
    mIoU  & 43.7  & 46.6  & \textbf{47.8} \\
\bottomrule
\end{tabular}

\end{small}
\end{center}
\vskip -0.1in
\end{table}
\footnotetext{Our reproduction}

\subsubsection{Comparison to Other Methods}

%\zijing{Maybe change the title into 'Comparison to few shot methods'.} 这段里包含了few shot和full两种

%\zijing{For methods with different settings, consider adding a split line in Table 2.}

% Our method outperforms DATUM, a one-shot method, in the day$\to$night, clear$\to$snow, and clear$\to$rain settings. Additionally, it surpasses DAFormer, which utilizes full target images, in the clear$\to$snow, clear$\to$rain, and clear$\to$fog settings. The success of our approach can be attributed to the generation of an augmented and diverse dataset that matches the scale of the source dataset. Although the synthesized data contains some noise, it still effectively aids the model in adapting to the new domain. However, in the day$\to$night and real$\to$game scenarios, our method, relying solely on text prompts, struggles to precisely capture the characteristics of the unseen target domain, resulting in performance inferior to that of DAFormer.

There exist other approaches that leverage unlabeled target domain data during the training process to achieve enhanced adaptation performance. Surprisingly, 
%\yang{as shown in which table? Please check and fix similar issues across the paper.}
as shown in Table \ref{maintab}, under the day$\to$night, clear$\to$snow, and clear$\to$rain settings, we even outperform DATUM, a one-shot method that utilizes a few target domain images, as well as DAFormer, which leverages full target domain images, in the clear$\to$snow, clear$\to$rain, and clear$\to$fog settings. This success is attributed to our ability to generate an augmented and diverse dataset that matches the scale of the source dataset. Although the synthesized data contains some noise, it still effectively aids the model in adapting to the new domain. In other benchmarks, such as day$\to$night and real$\to$game, our method achieves comparable performance. The slight performance gap in these scenarios can be attributed to the inherent challenge of relying solely on text prompts to precisely capture the characteristics of the unseen target domain, particularly in cases where the domain shift is more pronounced or complex.

% \zijing{Change the expression of this paragraph. These are all few-shot or uda methods, it's a surprise that we outperform them, and it's normal if we do not. Try to express like ``Under xxx and xxx benchmarks, we even outperform xxx method which use few shot target domain images and xxx which use all target domain images. In other benchmarks we also achieve comparable performance.'' If possible, list some reason why we still have some gap in day2night and real2game benchmarks.}

\subsection{Ablation Study}

% \zijing{
% More suggested ablation experiments on synthetic data generation:
% 1. The ratio of cropping patch. For example, for synthetic target domain, why not resize to a larger image and crop it into 4 pieces.
% 2. Why random cropping. For example, why not center cropping to generate intermediate domain.
% 3. Can we paste with a transparant ratio to create augmentated domain?
% 4. Can we train the model with patches instead of concanating the patching or pasting the patches?
% }

% \zijing{
% More suggested ablation experiments on progressive adaptation:
% 1. Can we train it in more adaptation ways, for example, (1) one stage, use all data. (2) two stages, first source pretrain, then intermediate and target finetune. (3) two stages, first source and intermediate pretrain, then target finetune. (4) Three stages, first source, then intermediate finetune, finally target finetune.
% 2. Can we apply more optimization strategies other than early stopping, for example, other loss decay methods.
% 3. How to determine the epoch number of early stopping. (If it purely depends on experiments, then it seems like your are tuning hyper-pareameter of epoch number based on testing set which in theory cannot be seen during training, especially in zero-shot setting where you don't have evaluation set. May need some investigation on finding the appropriate early stopping epochs.)
% }

\subsubsection{Intermediate Domain}

The intermediate domain is a key component of our method. It not only serves as an augmented domain bridging the source domain and the target domain but also enables the model to progressively adapt to the target domain. To investigate its importance, we conducted an ablation study by removing the intermediate domain and directly training the model on the source and synthetic target domains for 100 epochs. The results, as shown in Table \ref{mediumtab}, demonstrate the significant role of the intermediate domain in achieving effective adaptation. We point out that in the Clear →Snow setting the model is more vulnerable to noise in synthetic data and stopping earlier is better. As evidenced in Table \ref{earlystoptab}, our method achieves 48.4±1.0 mIoU with intermediate domain and early stop at 31st epoch (instead of 65th epoch by default), outperforming the setting without intermediate domain, which achieves 47.5±1.6 mIoU.  %\yang{not an improvement of 48? just achieve, also 48.4 seems like a incorrect number? which number I should look into in that table, I thought you use 65 by default.}

% \zijing{Abnormal results like clear2snow performance needs to be explained. For convinience, I recommand fixing the experiments to make the number at least lower (even by 0.1) than w/ Intermediate.}

\begin{table}[ht]
\caption{The impact of synthetic intermediate domain on our method’s final mIoU
under different settings.For every experiment, we use the same hyperparameters.}
\label{mediumtab}
\vspace{-8pt}
\begin{center}
\begin{small}
\begin{tabular}{lcc}
\toprule
          & \textnormal{w/ Intermediate} & \textnormal{w/o Intermediate} \\
\midrule
    Day →Night & \textbf{26.9}±0.8 & 24.1±1.6 \\
    Clear →Snow & 47.4±0.7 & \textbf{47.5}±1.6 \\
    Clear →Rain & \textbf{48.6}±0.8 & 46.1±1.0 \\
    Clear →Fog & \textbf{58.8}±0.7 & 58.2±2.0 \\
    Real →Game & \textbf{43.4}±0.4 & 42.3±0.4 \\
\bottomrule
\end{tabular}
\end{small}
\end{center}
\vskip -0.1in
\end{table}

\subsubsection{Random Cropping}
\label{sec:abl-random-cropping}

We conducted experiments on the clear$\to$rain setting with two distinct configurations. In the first setting (denoted as ``w/o randomness''), we avoided random cropping for constructing the intermediate domain. Instead, we extracted the center patch from all images to build the intermediate domain. In the second setting (denoted as ``high res''), we operated at the original resolution of 2048 $\times$ 1024. Here, the source image was divided into 8 patches of size 512 $\times$ 512, and four of these patches were randomly selected for editing to construct the intermediate domain. The results are presented in Table \ref{randomtab}. The first setting, which lacks data augmentation, demonstrated relatively poor performance. In contrast, the second setting exhibited performance comparable to SDGPA's default setting in Section \ref{sec:method}, as both share a similar level of randomness. However, it requires four times the computational resources of SDGPA's default setting, which is a significant drawback.

\begin{table}[ht]
\caption{A comparison of different cropping strategies.}
\label{randomtab}
\vspace{-8pt}
\begin{center}
\begin{small}

\begin{tabular}{cccc}
\toprule
          & w/o randomness & high res & SDGPA \\
\midrule
    mIoU  & 47.6±0.6 & 48.5±1.3 & \textbf{48.6}±0.8 \\
\bottomrule
\end{tabular}

\end{small}
\end{center}
\vskip -0.1in
\end{table}

\subsubsection{Training Strategy}
\label{sec:abl-training-strategy}

We conducted experiments on the clear$\to$rain setting with two distinct configurations. In the first setting (denoted as ``w/o progressive''), we trained the model without progressive training, utilizing all the data for 100 epochs. In the second setting (denoted as ``finetune''), we implemented an alternative progressive training strategy: the model was first pre-trained on the source domain for 100 epochs, followed by fine-tuning on the intermediate domain for 50 epochs, and finally fine-tuning on the target domain for another 50 epochs. The results are presented in Table \ref{progressivetab}. These results demonstrate that our proposed
%\yang{word original sometimes misleading, suggest change to our proposed; Also in table VI, last column first row, change keyword original to Ours}
progressive training strategy achieves the best adaptation performance.

\begin{table}[ht]
\caption{A comparison of different training strategies.}
\label{progressivetab}
\vspace{-8pt}
\begin{center}
\begin{small}

\begin{tabular}{cccc}
\toprule
          & w/o progressive & finetune & SDGPA \\
\midrule
    mIoU  & 46.4±0.6 & 46.0±1.3 & \textbf{48.6}±0.8 \\
\bottomrule
\end{tabular}

\end{small}
\end{center}
\vskip -0.1in
\end{table}

\subsubsection{Early stopping}

Early stopping is a widely used and effective technique in deep learning, particularly when dealing with noisy labels\cite{bai2021earlystopping}. Our synthetic intermediate domain and synthetic target domain inherently contain noise, making early stopping especially relevant. During the first stage of Progressive Adaptation, the segmentation model is sufficiently trained to achieve strong segmentation capabilities. In the second stage, the primary objective is to transfer this capability to the unseen target domain. Therefore, we can and should employ early stopping to prevent overfitting to noisy data. The impact of early stopping is detailed in Table \ref{earlystoptab}. Stopping too early (e.g., at the 31st epoch) may result in insufficient model adaptation, whereas stopping too late (e.g., at the 100th epoch) may lead to overfitting due to noise in the synthetic dataset. We find that stopping at an epoch where the model achieves a high mIoU on the validation set of the source domain (e.g., at the 65th epoch) is often beneficial.
%\zijing{Check the grammar.}

\begin{table}[ht]
\caption{The impact of early stopping on our method's final mIoU under different settings. For every experiment, we use the same hyperparameters. The number at the head of the table means which epoch we stop at during the second stage of Progressive Adaptation.}
\label{earlystoptab}
\vspace{-8pt}
\begin{center}
\begin{small}

\begin{tabular}{lccc}
\toprule
    Setting/Stop at & \textbf{31} & \textbf{65} & \textbf{100} \\
\midrule
    Day →Night & 25.2±0.9 & \textbf{26.9}±0.8 & 26.1±0.8 \\
    Clear →Snow & \textbf{48.4}±1.0 & 47.4±0.7 & 46.6±0.5 \\
    Clear →Rain & 48.3±0.8 & \textbf{48.6}±0.8 & 48.0±0.2 \\
    Clear →Fog & 57.9±1.8 & \textbf{58.8}±0.7 & 58.3±2.1 \\
    Real →Game & 42.8±1.0 & \textbf{43.4}±0.4 & 43.3±0.4 \\
\bottomrule
\end{tabular}

\end{small}
\end{center}
\vskip -0.1in
\end{table}

\subsubsection{Influence of Hyperparameters}

\begin{figure}[ht]
\vskip 0.2in
\begin{center}
\centerline{\includegraphics[width=\columnwidth]{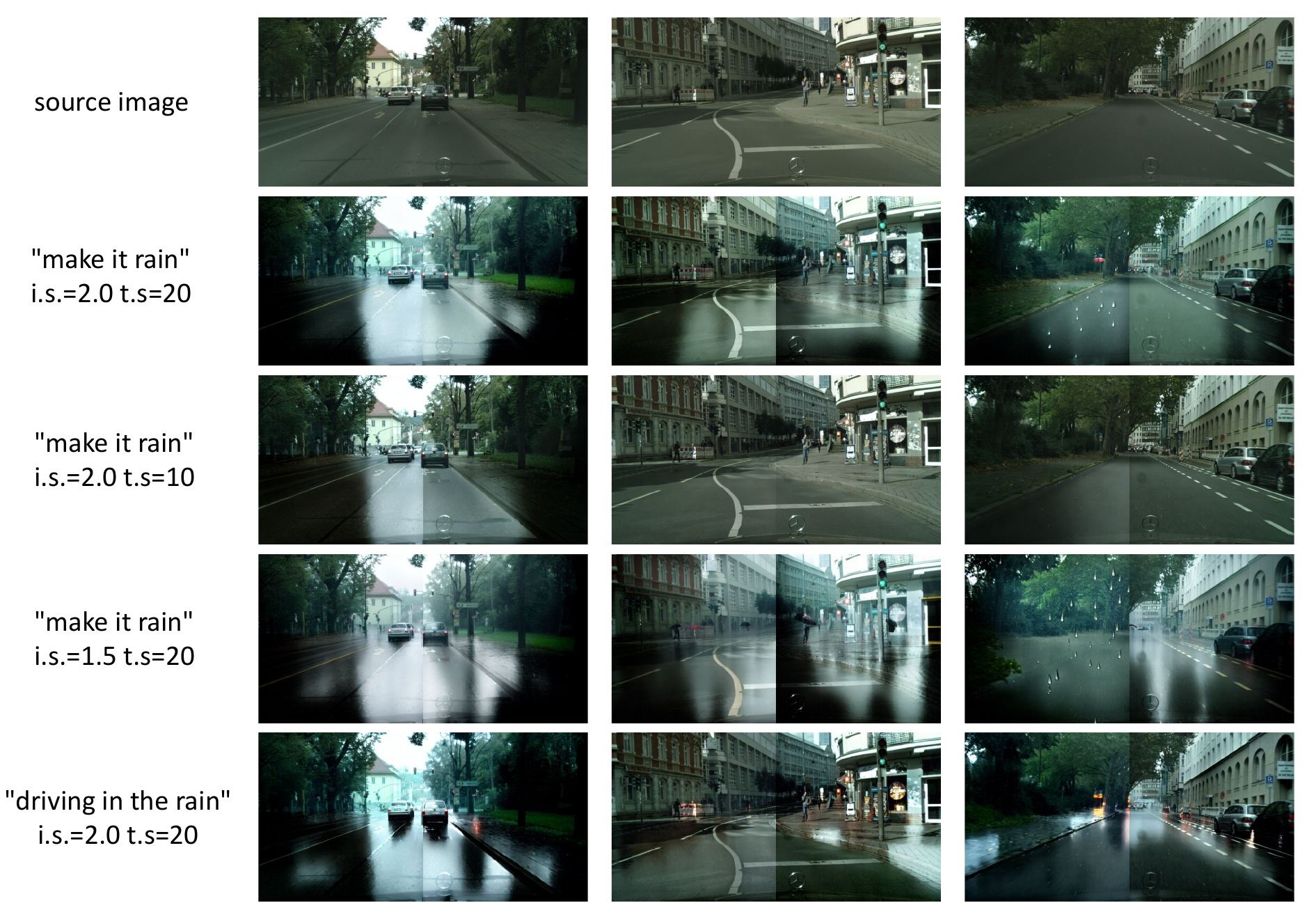}}
\caption{\textbf{Influence of Hyperparameters}. We selected three images under the clear$\to$rain setting as examples. The first image in each column is the original image from the source domain, and the remaining four images are synthetic target domain images obtained by using our method with different hyperparameters (as marked on the left side of the figure). i.s. is short for image guidance scale, and t.s. is short for text guidance scale. In row 2. we achieve a better tradeoff between image content and desired style.}
\label{hyperparamfig}
\end{center}
\vskip -0.2in
\end{figure}

When using InstructPix2Pix for image editing, the guidance scales for both the image and text significantly influence the quality of the generated images. Our goal is to retain as much information from the original image as possible while transforming its style, which inherently involves a tradeoff. Following InstructPix2pix's recommendation, we adjust both the image guidance scale and text guidance scale simultaneously. Simultaneously, to ensure a good tradeoff, we visually inspect some of generated images for style transfer quality and content preservation after each hyperparameter adjustment. Empirically, we observe that employing edit prompts included in InstructPix2Pix's training set often leads to better performance and facilitates achieving a more favorable tradeoff. A visual demonstration of influence of hyperparameters in the clear$\to$rain setting is illustrated in Figure \ref{hyperparamfig}.

% \yang{\subsubsection{Failure Analysis}}
% \yang{I suggest move failure case figs and discussion here, make it concise.}

\section{Conclusion}

This paper effectively addresses the zero shot domain adaptative semantic segmentation problem by leveraging the capabilities of a diffusion model based image editing model. We propose a novel synthetic data generation method that successfully constructs augmented intermediate domain and target domain. Furthermore, we introduce a progressive adaptation strategy that efficiently utilizes the noisy synthetic intermediate and target domains, enabling the model to better adapt to the unseen target domain while minimizing the impact of noise. Experimentally, our approach achieves state-of-the-art performance across five different settings. Extensive ablation experiments demonstrate the rationality and effectiveness of our method.

\section{Limitations}

\begin{figure}[ht]
\vskip 0.2in
\begin{center}
\centerline{\includegraphics[width=0.9\columnwidth]{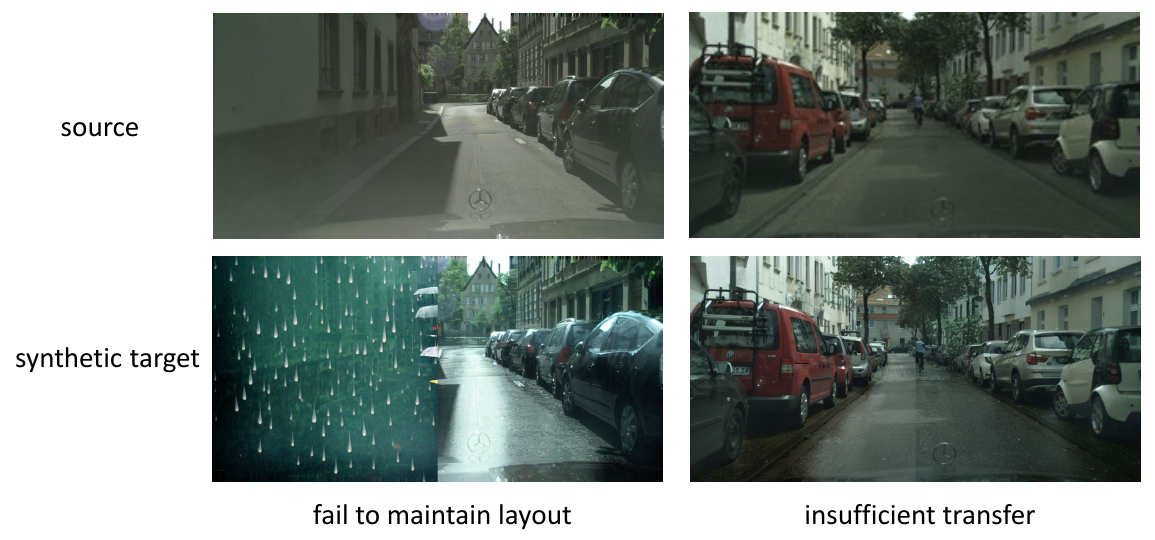}}
\caption{\textbf{Failure Cases}. We selected two images under the clear$\to$rain setting as examples. The synthetic target image in the first column fails to maintain the layout of the source image, while the synthetic target image in the second column fails to adequately transform into a rainy-day scenario.}
\label{limitfig}
\end{center}
\vskip -0.2in
\end{figure}

Our approach relies on the image editing model InstructPix2Pix's ability to retain image layout during style editing, but does not fully eliminate potential layout changes. 
%\zijing{Just say our method rely on the performance of pretrained diffusions, don't say that we have not tested other models.}
Despite our careful hyperparameter tuning, some patches from source images still undergo insufficient style transformation or excessive layout changes, as shown in Figure \ref{limitfig}. Although such failures are rare in the synthetic dataset, they can nevertheless degrade the segmentation model's performance. Our proposed method helps mitigate this impact, but cannot fully solve the problem. We hope that future research explores methods for filtering low-quality images or enhancing the segmentation model's robustness to address this limitation.
%\zijing{I suggest not mentioning the difficulty of tuning hyperparameters. Just say that failure images exist, which are rare but can harm the performance. Our proposed method help mitigate this impact but cannot fully solve the problem. We hope future studies investigate the method of filtering out the bad images or improve the robustness of the segmentation model and help solve this problem. Change the expression and make it seems like a not that serious problem.}

% Our approach employs InstructPix2Pix as the image editing model and does not explore the applicability of more models. Our method relies on the ability of the image editing model to maintain the layout of the image while editing its style, but it does not completely resolve potential layout changes. Our approach requires careful design and adjustment of hyperparameters; otherwise, it may lead to insufficient style transformation of the source image or excessive layout changes, as illustrated in Figure \ref{limitfig}. Although such failure cases are not numerous in the synthetic dataset, they can still harm the performance of the segmentation model. Our progressive model adaptation method can mitigate such noise in the dataset.

{\small
\bibliographystyle{IEEEtran}
\bibliography{main}
}

\end{document}